\documentclass[10pt,twocolumn,letterpaper]{article}

\usepackage{cvpr}              

\usepackage{multirow}

\usepackage[dvipsnames]{xcolor}

\definecolor{cvprblue}{rgb}{0.21,0.49,0.74}
\usepackage[pagebackref,breaklinks,colorlinks,citecolor=cvprblue]{hyperref}

\def\paperID{4689} 
\def\confName{CVPR}
\def\confYear{2024}

\usepackage{xspace}
\newcommand{\MethodName}{UniLSeg\xspace}

\title{Universal Segmentation at Arbitrary Granularity with Language Instruction}

\author{Yong Liu\textsuperscript{1}~,
        Cairong Zhang\textsuperscript{2}~,
        Yitong Wang\textsuperscript{2}~,
        Jiahao Wang\textsuperscript{3}~,
        Yujiu Yang\textsuperscript{1}~
        Yansong Tang\textsuperscript{1}\footnotemark[1]~\\
\textsuperscript{1}Tsinghua Shenzhen International Graduate School, Tsinghua University\\
\textsuperscript{2}ByteDance Inc.~
\textsuperscript{3}The University of Hong Kong\\
{\tt\small liuyong23@mails.tsinghua.edu.cn, tang.yansong@sz.tsinghua.edu.cn}
}

\begin{document}

\twocolumn[{
    \renewcommand\twocolumn[1][]{#1}
    \maketitle
    \begin{center}
        \includegraphics[width=1.0\linewidth]{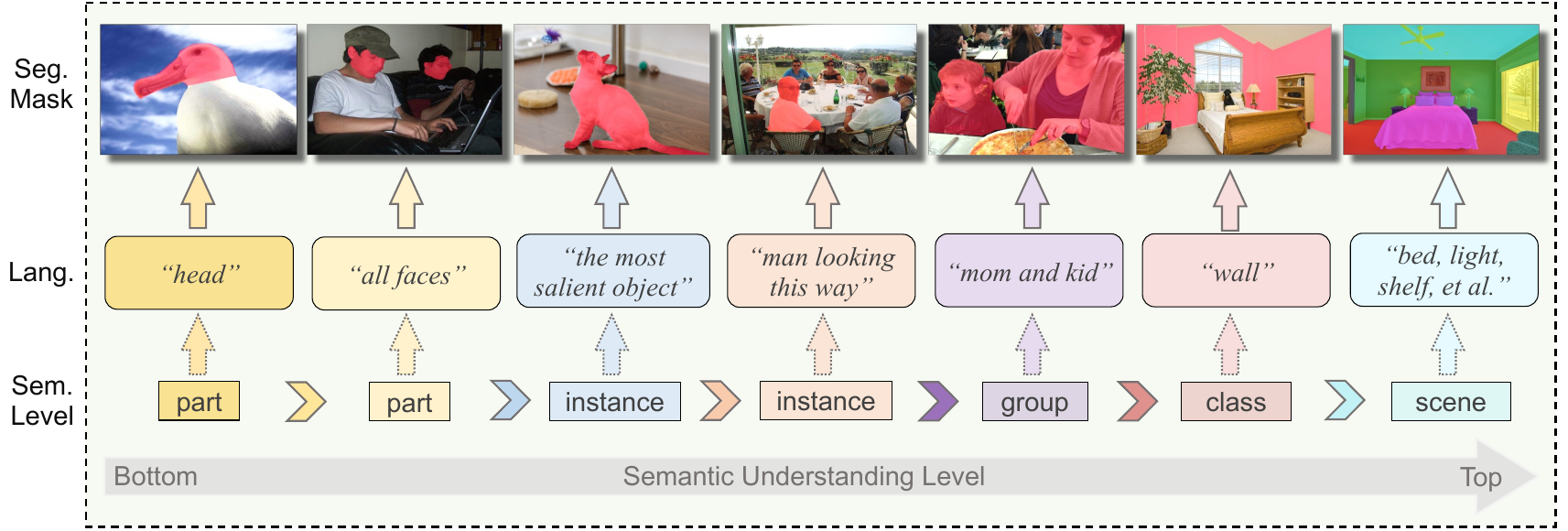}
        \vspace{-20pt}
        \captionof{figure}{Illustration of our \MethodName that is able to segment images at any granularity or semantic level with language as instructions. ``Seg. Mask", ``Lang.", and ``Sem. Level" denote the segmentation masks, corresponding language descriptions, and semantic levels, respectively. The segmentation masks are shown in red or other colors. \MethodName can take arbitrary text as input, whether it is a detailed long description of an object or a short category name.
        With flexible expressions indicating segmentation target, \MethodName achieves excellent performance on various semantic level, \eg, object part, single or multiple instances, and the whole scene.}
        \label{fig:teaser}
    \end{center}
}]

\footnotetext[1]{Corresponding author}

\begin{abstract}
\vspace{-10pt}
This paper aims to achieve universal segmentation of arbitrary semantic level.
Despite significant progress in recent years, specialist segmentation approaches are limited to specific tasks and data distribution. Retraining a new model for adaptation to new scenarios or settings takes expensive computation and time cost, which raises the demand for versatile and universal segmentation model that can cater to various granularity. 
Although some attempts have been made for unifying different segmentation tasks or generalization to various scenarios, limitations in the definition of paradigms and input-output spaces make it difficult for them to achieve accurate understanding of content at arbitrary granularity. 
To this end, we present \MethodName, a universal segmentation model that can perform segmentation at any semantic level with the guidance of language instructions. 
For training \MethodName, we reorganize a group of tasks from original diverse distributions into a unified data format, where images with texts describing segmentation targets as input and corresponding masks are output. Combined with a automatic annotation engine for utilizing numerous unlabeled data, \MethodName achieves excellent performance on various tasks and settings, surpassing both specialist and unified segmentation models. Code is available \href{https://github.com/yongliu20/UniLSeg}{here}.
\end{abstract}

\vspace{-7pt}
\section{Introduction}
Segmentation is one of the most important problem in computer vision, which aims to group meaningful regions and perform pixel-level understanding. Recent years have witnessed great progress in the development of various segmentation tasks such as semantic segmentation~\cite{fcn, mask2former, gkc, scan}, interactive segmentation~\cite{sam, interactive1, gsfm, qdmn}, salient object segmentation~\cite{icon, rcsbnet}, and referring segmentation~\cite{lavt, soc}.

Although many excellent works have emerged, they tend to be specialist approaches for specific segmentation tasks, making it difficult for them to address complex and diverse segmentation scenarios. When adapting to novel settings or semantics, new models need to be designed and trained on data of corresponding distribution, which leads to significant data and computation cost. Therefore, it is greatly promising to achieve versatile and universal segmentation that can cater to various semantic levels and settings. Nevertheless, due to the diverse distribution of data and the complexities of input-output space, designing and training such a model presents a significant challenge.

Recently, some works~\cite{sam, seem, uninext, seggpt} have attempted to propose unified paradigm for multiple segmentation tasks or generalization to various scenarios. Among them, SAM~\cite{sam} and SEEM~\cite{seem} propose to take point as the basis for indicating segmentation targets and have achieved impressive performance and generalization ability. However, such point-based interaction paradigm tends to produce over-dispersed and unexpected segmentation results due to the lack of semantic concept awareness. Besides, the information contained within the ``point" is insufficient to guide the model in executing generic segmentation across multiple semantic levels. Unlike SAM, UNINEXT~\cite{uninext} focuses on object-centric segmentation and adopt prompt generation to standardize the input space. Despite achieving excellent results, the emphasis on instance makes it difficult to achieve understanding of any granularity, such as fine-grained part segmentation and coarse-grained scene understanding. With visual in-context learning, Painter~\cite{painter} and SegGPT~\cite{seggpt} realize flexible target-aware segmentation. But the unification of input-output example form of different tasks  has brought up new challenges and greatly limits the application scenarios.

Therefore, it leads to a natural question: \textit{is there a unified paradigm that can conveniently interact with the model for universal segmentation at any granularity and has good scalability?} 
The answer we proffer is language. As the most important tool for humanity, language can flexibly express the objects of reality and the laws of thought. 
Like communication between human beings, language has become one of the most important means of communication between humans and machines. It has the capability to provide information at various levels of detail, effectively guiding the model in accomplishing the desired tasks. Such versatile and informative prompt is in line with the above requirements for unified segmentation instruction.
With this insight, we study a series of tasks to validate our ideas. The explored tasks contain referring image segmentation (RIS)~\cite{lavt, cris}, semantic segmentation (SS)~\cite{fcn, mask2former}, salient object detection (SOD)~\cite{sod1, sod2}, part segmentation (PS)~\cite{semanticsam, partimagenet}, referring video object segmentation (RVOS)~\cite{soc, urvos}, and open-vocabulary segmentation (OVS)~\cite{gkc, Simbaseline}. We reorganize these tasks from original diverse distributions into a unified data format, where images with texts describing segmentation targets as input and corresponding masks are output. Benefiting from such flexible and unified design, segmentation model can be jointly trained on different tasks to learn the connections between language instructions and visual concepts. 

In addition, to promote the the model's understanding of high-level language instructions, we present a fully aligned framework called \MethodName. The core design philosophy of \MethodName lies in conducting extensive visual-linguistic interaction to identify and segment the targets within multi-modal joint space.
As shown in \Cref{fig:teaser}, with generic representations learned from numerous data of various tasks, our \MethodName  is able to comprehend diverse language expressions that indicate segmentation targets at varying semantic levels.
Furthermore, to exploit the large-scale unlabelled and weakly supervised data, we propose an automatic annotation engine to generate pseudo caption-mask pairs for assisted training. 
Extensive experiments on a group of benchmarks demonstrate the powerful  segmentation ability of our \MethodName, \eg, it overpasses specialist models and unified competitors by about 12\% and 7\% on G-Ref~\cite{grefcoco} validation set.

Our contributions can be summarized as follows:
\begin{itemize}
    \item We present \MethodName, a generic segmentation model that fully integrates visual concept with textual guidance. With language instructions as universal prompt, \MethodName can be jointly trained on multiple tasks to learn the connections between diverse textual descriptions and visual content, achieving universal segmentation at arbitrary semantic granularity.
    \item Our \MethodName achieves superior performance on a group of challenging benchmarks from various tasks, which benefits from the language-based paradigm definition and additional automatic annotation engine for large-scale unlabeled and weakly-labeled data.
\end{itemize}


\section{Related Work}
\paragraph{Unified Large Segmentation Model}
Some recent works have been devoted to exploring unified and robust large segmentation models. Among them, SAM~\cite{sam} has received great attention for its powerful and generalizable segmentation ability. SAM proposes to leverage point-based prompts for indicating target regions. A series of follow-up works~\cite{trackanything, semanticsam, ram, sca} have built on SAM, applying it to a variety of tasks and achieving outstanding performance. However, random point sampling and point-based interaction may lead to over-dispersed segmentation masks and unawareness of high-level semantics. In addition, Painter~\cite{painter} and SegGPT~\cite{seggpt} leverage visual in-context learning and reorganize the output space of different tasks for unified prediction. UNINEXT~\cite{uninext} and SEEM~\cite{seem} design unified segmentation decoder for adapting different task prompts.
Although these works have achieved excellent performance, limitations in the definition of paradigms and input-output space make it difficult for them to perform segmentation at any semantic granularity.

\begin{figure*}[t]
    \centering
    \includegraphics[width=\textwidth]{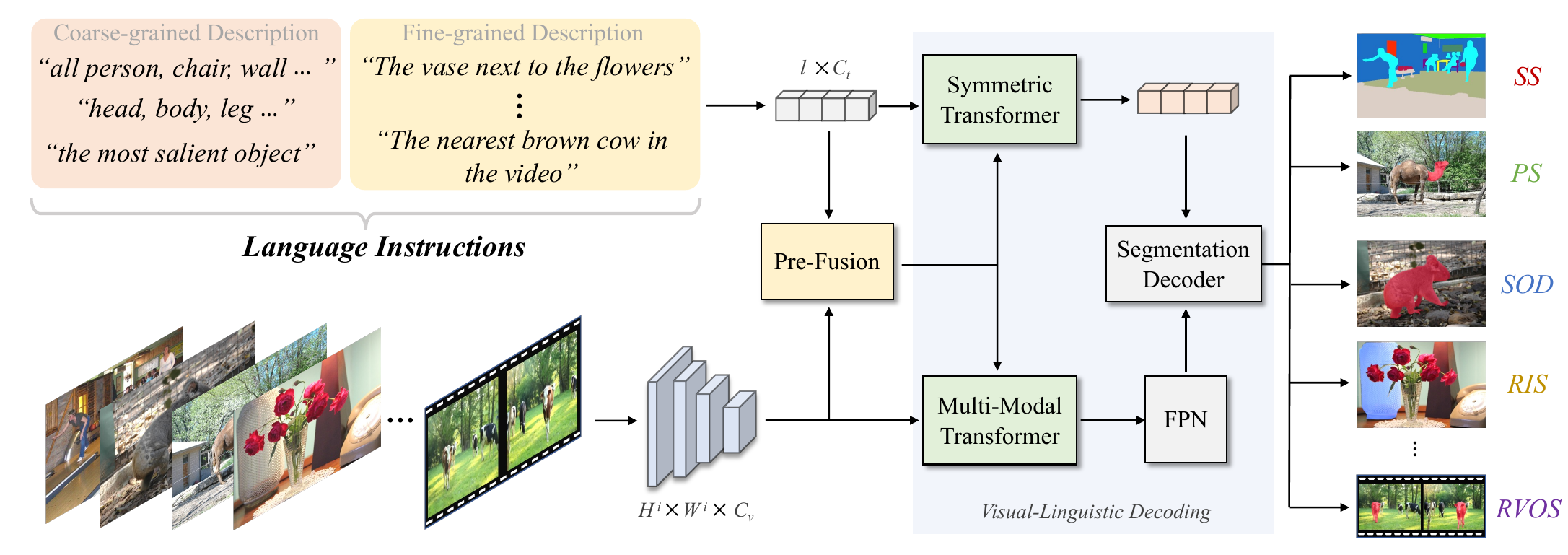}
    \vspace{-10pt}
    \caption{Pipeline of our \MethodName. It takes both images and corresponding language prompt as input.
    With versatile language descriptions indicating segmentation targets and full visual-linguistic interactions, \MethodName can perform segmentation at any semantic granularity and tackle various tasks such as semantic segmentation (SS), part segmentation (PS), salient object detection (SOD), open-vocabulary segmentation (OVS), referring image (RIS) and video object segmentation (RVOS).}
    \label{fig:pipeline}
    \vspace{-10pt}
\end{figure*}

\vspace{-5pt}
\paragraph{Language-Guided Segmentation}
As the most representative language-guided segmentation task, referring image segmentation aims to perform pixel-level visual-linguistic alignment for input images and given descriptions.
The pioneer works~\cite{early-work1,early-work2} extracts image and language features respectively and concatenates them to form the multi-modal features. 
The subsequent approaches generally can be divided into two categories.
The first idea~\cite{mattnet, cmpc, busnet} is utilizing the internal structure of the text to help identify target objects.
However, this approach does not model well-aligned cross-modal joint space, and the pipeline tends to be complex.
The other idea~\cite{step, cmsa, kwn, brinet, lts, efn, vlt, restr, lavt, cris} is to model the cross-modal relations between image and language by various attention operations.
Following the attention alignment idea, there are many follow-up works such as LAVT~\cite{lavt}, GRES~\cite{gres}, and PolyFormer~\cite{polyformer}. 
In addition to referring segmentation, semantic segmentation can also be taken as language-guided task and the category names can be viewed as short and rough text descriptions~\cite{lseg, Simbaseline, gkc, openseg}.
Different from  the detailed descriptions in referring segmentation, prior guidance provided by class name is coarse-grained, which poses challenges in multi-modal interaction. Even with the help of additional vision-language models such as CLIP~\cite{CLIP}, the model performance is still unsatisfactory.


\section{Method}

\subsection{Pipeline}

\textbf{Overview: }\Cref{fig:pipeline} shows the pipeline of our \MethodName. The core design philosophy of \MethodName lies in performing extensive visual-linguistic interaction, which is compatible with the language-based unified paradigm.  Specifically, it takes both images and corresponding language prompt as input. By perceiving segmentation target in cross-modal joint space and activating corresponding response, \MethodName achieves universal segmentation of arbitrary semantic granularity, represented by a group of tasks.
We will elaborate it in the following sections.

\vspace{5pt}
\noindent \textbf{Encoding Process:}
For the input image $I \in \mathbb{R}^{H\times W\times 3}$, we utilize a pyramidal vision encoder~\cite{swin} to extract the hierarchical vision feature $f_v^i \in \mathbb{R}^{H^i\times W^i\times C_v^i}$, $i \in$ [1,2,3,4].
Here $H^i$ and $W^i$ denote the height and width of $i\text{-}th$ scale feature map, respectively. $C_v$ denotes the channel dimension of visual features.

For the input language prompt $L \in \mathbb{R}^{l}$, we take the transformer-based language encoder~\cite{CLIP} to encode it to a word embedding $f_w \in \mathbb{R}^{l\times C_t}$ and an overall sentence embedding $f_s \in \mathbb{R}^{1\times C_t}$, where $l$ is the length of the input language expression.
The word embedding $f_w$ contains fine-grained guidance information.
The sentence embedding $f_s$, on the other hand, expresses the general characteristics of segmentation targets.
Joint utilization of $f_w$ and $f_s$ contributes to different tasks.

\vspace{5pt}
\noindent \textbf{Pre-Fusion:}
Pre-Fusion aims to incorporate language guidance into visual features and 
roughly highlight target areas, which helps to mitigate the impact of background noise to visual-linguistic joint space. 
During our exploration, we have discovered that this module does not require a complex design.
Implemented with simple multi-head cross-attention, this part can achieve desired activation effect.
Specifically, Pre-Fusion takes the word feature $f_w$ and hierarchical vision feature $f_v^i$, $i \in$ [2,3,4] as input. Here take $i\text{-}th$ scale as an example for illustration. Since the purpose of this structure is to elicit potential response within visual content, we take the visual feature $f_v^i$ as the Query and the word embedding $f_w$ as the Key and Value. The process can be formulated as:
\begin{equation}
        \begin{split}
            &f_c^i = softmax(\frac{G_q(f_v^i)^TG_k(f_w)}{\sqrt{C^i}})G_v(f_w)^T,
        \end{split}
\end{equation}
where $G_q$, $G_k$, $G_v$ are projection functions that transfer the input features to the corresponding space. $C$ denotes channel dimension of joint embedding space. $f_c$ is activated visual feature used for subsequent visual-linguistic decoding.

\vspace{5pt}
\noindent \textbf{Visual-Linguistic Decoding:}
We design a two-stream decoding structure to fully utilize the guidance from language instructions and align cross-modal joint space. 
Specifically, in the vision path, the visual features are sent to a Multi-Modal Transformer for learning intra- and inter-modality connections.
In the language path, we incorporate the activated visual context into linguistic prompt embedding to generate content-aware prompts. This approach enhances the alignment between visual and linguistic spaces and effectively reduces cross-modal domain gaps. Below we present the details of these two paths.

\textit{Vision Path:}
This path consists of a Multi-Modal Transformer and a FPN~\cite{fpn}. 
Due to the limited alignment capacity of Pre-Fusion, the multi-modal transformer is used to perform sufficient intra- and inter-modality interactions.
It is mainly composed of multi-modal self-attention and multi-modal cross-attention operation.
Taking the $i\text{-}th$ level visual feature $f_c^i$ as an example, we first flatten it along the spatial dimension and add fixed positional embeddings~\cite{detr} to it.
After that, the flattened visual tokens are concatenated together with word embeddings $f_w$ to form multi-modal tokens $f_m$.
Then multi-head self-attention is applied to extract relevant information between them.
Self-attention allows the model to excavate information within respective modalities while modeling the visual-linguistic joint  space.
Note that only the output vision tokens  are used for subsequent process, and the word tokens are discarded. 
This process can be formulated as:
\begin{align}
           &f^i_m = flatten(f^i_c) + Pos.,\\
           &f^i_m = Concat(f^i_m, f_w),\\
           &f^i_b = MHSA(f^i_m)[:H^iW^i],\\
           &f^i_b = LN(f^i_b) + f^i_c,
\end{align}
where $MHSA$ and $LN$ denote multi-head self-attention and layer normalization~\cite{ln} operation, respectively.
$Pos$ is the fixed sinusoidal positional embeddings.
After that, we further leverage the output vision tokens $f^i_b$ as Query and the word embedding $f_w$ as Key and Value for multi-head cross-attention, benefiting the location of target regions.
Finally, a FPN-like~\cite{fpn} structure is utilized to integrate the aligned visual features of all scales.

\textit{Language Path:}
Inspired by prompt learning~\cite{learned_prompt, learned_prompt2, denseclip}, we also utilize a language instruction updating strategy to adjust linguistic space with visual content.
For simplicity and elegance of structure, we still take attention operation to achieve this goal. 
Actually, the process of language path is symmetric with the multi-modal transformer in vision path, \ie, symmetric transformer. It first takes cross-attention with sentence-level textual embedding $f_s$ as Query and the activated visual features $f_c$ as Key and Value. After that, self-attention is used to fully integrate initial language prompt with the content-aware one.

Finally, the activated visual features and content-aware linguistic embedding are combined to generate response map by similarity calculation, \ie, matrix multiplication. With bi-linear interpolation and binarization, the model generates the output mask.




\begin{table}
    \centering
    \small
    \renewcommand\arraystretch{1.1}
    \vspace{-8pt}
    \caption{Task-specific language prompt designs.}
    \setlength\tabcolsep{3.0pt}
    \label{tab:prompt}
    \begin{tabular}{l|c}
    \toprule
    Task  &Prompt Template  \\
    \hline
       Referring Image Segmentation & natural caption \\
       Referring Video Object Segmentation & natural caption \\
       Salient Object Detection & ``the most salient object" \\
       Semantic Segmentation & ``all  \{\}" \\
       Open-Vocabulary Segmentation & ``all  \{\}" \\
       Part Segmentation & ``all \{\}" \\
       \bottomrule
       
    \end{tabular}
    \vspace{-15pt}
\end{table}
\subsection{Task-Specific Prompt Design}
As the most flexible prompt, language descriptions can be reorganized to fit different goals. For tasks discussed in this paper, we design specific prompt templates for them and the summary is shown in \Cref{tab:prompt}. 

Referring image segmentation and referring video object segmentation aim to segment objects from images and videos based on a given language description. Since the related expressions already exist in these task, we directly leverage them as the language prompt. 
Semantic segmentation and open-vocabulary segmentation can be reformulated as language-guided paradigm by replacing output layers with computing the similarity between visual and linguistic embeddings. An intuitive approach is to take the category names as short textual expression. However, we find that this may potentially create semantic conflicts with long texts from referring segmentation and affects the effectiveness of joint training. To better combine data from different distributions, we design the language prompt of such category-based tasks to ``all \{\}", where \{\} denotes the target category name. 
For salient object detection, we directly utilize ``the most salient object" as the input template.

\subsection{Automatic Annotation for Unlabeled Data}
To train our \MethodName and make it capable of universal segmentation of arbitrary semantic granularity, we collect and reorganize an amount of supervised training data from available benchmarks. 
Specifically, these ``supervised data" come from RefCOCO~\cite{refcoco}, GRefCOCO~\cite{gres}, COCO-Stuff~\cite{coco}, Ref-YouTubeVOS~\cite{urvos}, PartImageNet~\cite{partimagenet}, LIP~\cite{lip}, ECSSD~\cite{ECSSD}, and DUTS~\cite{duts-tr}. 
We convert them to unified format, \ie, the triplet of image, mask, and corresponding language caption. Captions for data of different tasks are defined according to the template described above.

In addition to these supervised data, we also try to leverage the numerous unlabeled images and weakly annotated data. In particular, we classify these weakly supervised or unlabeled data into three categories: box-labeled, mask-labeled, and unlabeled. 
Since we take the language as unified prompt, we design an automatic annotation engine to generate desired caption-mask pairs and filter out potential noise.
1) For box-labeled data that is mainly collected from Object365~\cite{obj365} and RefCOCO~\cite{refcoco}, we crop related sub-images with annotated bounding boxes and then leverage SAM~\cite{sam} and BLIP~\cite{blip} to generate pseudo masks and captions.
2) For mask-based data sampled from SA-1B dataset~\cite{sam}, a naive method is captioning each annotated mask region with image-caption models. However, we find that this would lead to category redundancy and severe mismatches between different masks of the same class. 
To address this problem,  we abandon the approach that labeling text based on the existing masks, and instead re-label both the masks and text from scratch. Specifically, we first leverage RAM~\cite{ram} to tag any common categories in images. With all categories exist in image as input, Grounding DINO~\cite{groundingdino} is introduced to detect object regions related to each class tag. Then, with bounding boxes generated by Grounding DINO as box prompt, we use the huge version of SAM~\cite{sam} to generate fine-grained masks for all potential categories.
3) For unlabeled data, \eg, images from ImageNet~\cite{imagenet}, we first use BLIP~\cite{blip} to generate natural captions for each image.  Then a referring segmentation model pre-trained on referring datasets~\cite{refcoco} is applied to recognize and segment the object related to the caption, obtaining the desired mask-caption pairs.

To remove undesired low-quality triplets and reduce the impact of annotation noise exist in pseudo-labeled data, we first leverage CLIP~\cite{CLIP} to calculate the matching score of each caption with corresponding mask region. By filtering triplets with excessively low matching score, the quality of these pseudo-labeled data is improved. Besides, we incorporate the hide-and-seek strategy~\cite{kmn} during the training process to effectively alleviate the detrimental effects of inaccurate pseudo labels. The patch hiding probability is 0.2.

\section{Experiment}

\subsection{Implementation Details.}
We take the text encoder of CLIP ViT-B/16~\cite{CLIP} and Swin Transformer~\cite{swin} pre-trained on ImageNet~\cite{imagenet} as our encoder in default.
Both language and vision encoder are initialized using the official pre-trained weights. The rest of weights in our model are randomly initialized.
The input images are resized to 480 × 480 by default and no data augmentation technique is applied.
We adopt a two-stage pre-training strategy to fully utilize data of different distributions.
For the first stage, we train the model on 192 Tesla V100 GPUs with the batch size of 8 for each GPU. Training data of the first stage is sampled from SA-1B dataset~\cite{sam}. The learning rate is set to 5e-5 and the epoch is set to 5.
For the second stage, the training data consist of supervised data collected from a group of benchmarks and remain pseudo labeled data. We train the model for 15 epochs in this stage with  learning rate of 1e-4. The learning rate is decayed by 0.1 after 10-$th$ epoch.
Besides, we also set a smaller learning rate for the visual backbone and the scaling factor is 0.1.
The model is optimized with the combination of  cross-entropy loss and Dice loss~\cite{dice-loss} with the Adam~\cite{adam} optimizer.
For convenience, we randomly sample one expression for each object within one iteration.
During inference, the output mask is upsampled to the size of the input image by bi-linear interpolation. We binarize the prediction masks by the threshold of 0.5 and do not utilize other post-process operations.

\subsection{Main Results}
We evaluate our \MethodName on 6 tasks with corresponding evaluation metrics. The inference benchmarks contain RefCOCO~\cite{refcoco}, RefCOCO+~\cite{refcoco}, G-Ref~\cite{grefcoco}, Ref-YouTubeVOS~\cite{urvos}, ADE20K~\cite{ade20k}, Pascal Context~\cite{pascal-voc}, PartImageNet~\cite{partimagenet}, ECSSD~\cite{ECSSD}, SOD~\cite{sod1}, and Pascal-S~\cite{pascal}. 
Note that we report the performance of two versions of \MethodName, \ie, \MethodName-20 and \MethodName-100, which differ in sampling proportion of the SA-1B dataset during pre-training.
The results are presented below.

\begin{table*}[ht]
    \centering
    \small
    \caption{Comparison with state-of-the-art methods in terms of oIoU on the popular referring image segmentation benchmarks RefCOCO~\cite{refcoco}, RefCOCO+~\cite{refcoco}, and G-Ref~\cite{grefcoco}. We compare our \MethodName with both proprietary approaches of RIS field and strong large segmentation models on both validation and test split.  ``-" represents that the result is not provided.}
    \setlength{\tabcolsep}{3.8mm}
    {\begin{tabular}{l|l|c|c|c|c|c|c|c|c}
      \hline
      \multirow{2}{*}{Method} &
      Vision &
      \multicolumn{3}{c|}{RefCOCO}  & \multicolumn{3}{c|}{RefCOCO+} & \multicolumn{2}{c}{G-Ref} \\
      \cline{3-10}
                                    & Backbone & val   & test A & test B & val & test A & test B & val & test   \\
      \hline
      \multicolumn{10}{c}{\textit{Proprietary Methods}}   \\
      \hline
      LSCM~\cite{lscm} & ResNet101 & 61.47 & 64.99 & 59.55 & 49.34 & 53.12 & 43.50 & -     & -\\
      CMPC+~\cite{cmpc+}      & ResNet101 & 62.47 & 65.08 & 60.82 & 50.25 & 54.04 & 43.47 & -     & -   \\
      MCN~\cite{mcn}       & DarkNet53 & 62.44 & 64.20 & 59.71 & 50.62 & 54.99 & 44.69 & 49.22 & 49.40 \\
      EFN~\cite{efn}                & ResNet101 & 62.76 & 65.69 & 59.67 & 51.50 & 55.24 & 43.01 & - & - \\
      BUSNet~\cite{busnet}          & ResNet101 & 63.27 & 66.41 & 61.39 & 51.76 & 56.87 & 44.13 & - & - \\
      CGAN~\cite{cgan}    & DarkNet53 & 64.86 & 68.04 & 62.07 & 51.03 & 55.51 & 44.06 & 51.01 & 51.69 \\
      LTS~\cite{lts}   & DarkNet53 & 65.43 & 67.76 & 63.08 & 54.21 & 58.32 & 48.02 & 54.40 & 54.25\\
      VLT~\cite{vlt}      & DarkNet53 & 65.65 & 68.29 & 62.73 & 55.50 & 59.20 & 49.36 & 52.99 & 56.65 \\
      ResTR~\cite{restr} & ViT-B & 67.22 & 69.30 & 64.45 & 55.78 & 60.44 & 48.27 & 54.48 & - \\
      CRIS~\cite{cris} & ResNet50 &69.52 &72.72 &64.70 &61.39 &67.10 &52.48 & 59.35 & 59.39 \\
      CRIS~\cite{cris} & ResNet101 &70.47 &73.18 &66.10 &62.27 &68.08 &53.68 & 59.87 & 60.36 \\
      LAVT~\cite{lavt} & Swin-B & 72.73 & 75.82 & 68.79 & 62.14 & 68.38 & 55.10 & 61.24 & 62.09 \\
      GRES~\cite{gres} & Swin-B & 73.82 & 76.48 & 70.18 & 66.04 & 71.02 & 57.65 & 65.00 & 65.97 \\
      PolyFormer~\cite{polyformer} & Swin-B & 74.82 & 76.64 & 71.06 & 67.64 & 72.89 & 59.33 & 67.76 & 69.05 \\
      
      \hline
      \multicolumn{10}{c}{\textit{Unified Segmentation Models}}   \\
      \hline
      SEEM~\cite{seem} & FocalT & - & - & - & - & - & - & 65.60 & - \\
      UniNext~\cite{uninext} &ConvNext-L & 80.32 & 82.61 & 77.76 & 70.04 & 74.91 & 62.57 & 73.41 & 73.68  \\
      \hline
      \MethodName-20 & Swin-B &80.52 & 81.83 &78.43 &72.70 &77.02 &66.99 &78.41 &79.47  \\
      \MethodName-100 & Swin-B &\textbf{81.74} & \textbf{83.17} &\textbf{79.85} &\textbf{73.18} & \textbf{78.29} & \textbf{68.15} &\textbf{79.27} & \textbf{80.54}  \\
      \hline
   \end{tabular}}
   
   \label{table:ris}
\end{table*}
\vspace{5pt}
\noindent \textbf{Referring Image Segmentation:}
Referring image segmentation task is the most appropriate measure for language-based segmentation because of the diversity and flexibility of the captions in this task. We compare our method with both proprietary approaches~\cite{lavt, cris, gres, polyformer} in RIS field and large unified segmentation models~\cite{seem, uninext}. The results are shown in \Cref{table:ris}. It can be seen that \MethodName surpasses all existing methods by a significant margin, \eg, 79.27 \textit{vs} 73.41 on G-Ref~\cite{grefcoco} validation set and 73.18 \textit{vs} 70.04 on RefCOCO+~\cite{refcoco} validation set.

\vspace{5pt}
\noindent \textbf{Referring Video Object Segmentation:}
\begin{table}[t]
   \centering
   \small
  \renewcommand\arraystretch{1.1}
  \setlength{\tabcolsep}{14.5pt}
  \caption{Results of referring video object segmentation on Ref-YouTubeVOS validation set.}
   \begin{tabular}{l|c|c|c}
      \hline
      Method & $\mathcal{J\&F}$ & $\mathcal{J}$  & $\mathcal{F}$\\
      \hline
      URVOS~\cite{urvos} & 47.2 & 45.3 & 49.2\\
      LBDT-4~\cite{lbdt} & 49.4 & 48.2 & 50.6\\
      MTTR~\cite{mttr} & 55.3 & 54.0 & 56.6\\
      VLT~\cite{vlt} &63.8 &61.9 &65.6 \\
      ReferFormer~\cite{referformer} &59.4	&58.0	&60.8\\
      SOC~\cite{soc} & 62.4 &61.1 &63.7   \\
      \hline
      \MethodName-20 &64.1 &61.9 &66.3\\
      \MethodName-100 &\textbf{64.9} &\textbf{62.8} &\textbf{67.0}\\
      \hline
   \end{tabular}
   \label{tab:rvos}
   \vspace{-5pt}
\end{table}
Results shown in \Cref{tab:rvos} demonstrate the effectiveness of our \MethodName in language-guided video segmentation. \MethodName handles videos in a frame-by-frame manner.
Actually, such performance stems in part from the lack of consideration of inter-frame relationships in the current RVOS benchmarks. However, robust image-level understanding ability also indicates the potential for extending our \MethodName to corresponding video-level model.

\vspace{5pt}
\noindent \textbf{Semantic Segmentation:}
Expressions in referring segmentation tend to be long texts that describes an object in detail.
To prove \MethodName has the ability to understand diverse linguistic descriptions and perform high-level scene understanding, we test it on semantic segmentation task under both out-vocabulary and in-vocabulary  settings with the coarse-grained category name as language instructions.

\begin{table}
    \centering
    \small
    \renewcommand\arraystretch{1.1}
    \caption{Results of open-vocabulary semantic segmentation with mIoU as metric. LMM denotes large multi-modal model.}
    \setlength\tabcolsep{4.5pt}
    \label{tab:ovs}
    \begin{tabular}{l|c|cc}
    \hline
    Method  &Additional LMM &ADE20K-150 &PC-59  \\
    \hline
       LSeg+~\cite{lseg} &$\times$ &13.0 &36.0\\
       SimSeg~\cite{Simbaseline} &\checkmark &20.5  &47.7 \\
       OpenSeg~\cite{openseg} &\checkmark  &21.1  &42.1 \\
       GKC~\cite{gkc} &$\times$ &18.8  &45.2 \\
       MaskCLIP~\cite{maskclip}  &\checkmark &23.7 &45.9 \\
       OVSeg~\cite{mask-adaptedclip} &\checkmark &24.8 &53.3 \\
       \hline
       \MethodName-20 &$\times$ &27.6  &54.3 \\
       \MethodName-100 &$\times$ &\textbf{29.5}  &\textbf{56.7} \\
       \hline
    \end{tabular}
    \vspace{-5pt}
\end{table}
\textit{Open-Vocabulary Setting:}
we directly take the pretrained \MethodName for open-vocabulary inference on ADE20K-150~\cite{ade20k} and Pascal Context-59~\cite{pascal-voc} datasets. From \Cref{tab:ovs} we can see that our \MethodName achieves excellent performance for open-vocabulary settings. It is worth noting that \MethodName even outperforms the methods utilizing additional large multi-modal models such as ALIGN~\cite{align} and CLIP~\cite{CLIP} for region classification.

\textit{In-Vocabulary Setting:}
we finetune the \MethodName on ADE20K~\cite{ade20k} dataset to evaluate it for the in-vocabulary setting. The results are shown in \Cref{tab:ss}. Although our approach  performs worse than the specialist models due to its uncustomized design about this task, it surpasses previous unified model that utilizes visual information for guidance, \textit{e.g.}, SegGPT~\cite{seggpt}, by a remarkable margin. 
\begin{table}
    \centering
    \small
    \renewcommand\arraystretch{1.1}
    \caption{Results of semantic segmentation with mIoU as metric.}
    \setlength\tabcolsep{17.2pt}
    \label{tab:ss}
    \begin{tabular}{lc}
    \hline
    Method  &ADE20K   \\
    \hline
       FCN+~\cite{fcn} &29.4 \\
       DeepLabV3+~\cite{deeplabv3}  &44.1   \\
       RefineNet~\cite{refinenet}  &40.7  \\
       SegFormer~\cite{segformer} &51.1  \\
       MaskCLIP~\cite{maskformer} &51.1 \\
       SegGPT~\cite{seggpt} &39.9  \\
       \hline
       \MethodName-20 &45.2 \\
       \MethodName-100 &49.5 \\
       \hline
    \end{tabular}
\end{table}

\vspace{5pt}
\noindent \textbf{Salient Object Detection:}
With the intuitive prompt ``the most salient object", our \MethodName achieves the best performance on all popular salient object detection benchmarks. \Cref{tab:sod} shows the corresponding comparison with $F_{mean}$ as metric. The larger is better. 
\begin{table}[t]
   \centering
   \small
  \renewcommand\arraystretch{1.1}
  \setlength{\tabcolsep}{11.5pt}
  \caption{Comparisons with salient object detection methods.}
   \begin{tabular}{l|c|c|c}
      \hline
      Method & ECSSD & SOD  & PASCAL-S\\
      \hline
      F3Net~\cite{f3net} & 0.912 & 0.775 & 0.816\\
      MINET~\cite{minet} & 0.911 & - & 0.809\\
      GateNet~\cite{gatenet} &0.894 &- &0.797 \\
      ICON~\cite{icon} & 0.936 &0.802 &0.854   \\
      MFABA~\cite{mfaba} & 0.935 &- &0.857   \\
      RCSBNet~\cite{rcsbnet} & 0.927 &- &0.842   \\
      \hline
      \MethodName-20 &0.954 &0.857 &0.881\\
      \MethodName-100 &\textbf{0.961} &\textbf{0.863} &\textbf{0.889}\\
      \hline
   \end{tabular}
   \label{tab:sod}
   \vspace{-5pt}
\end{table}

\vspace{5pt}
\noindent \textbf{Part Segmentation:}
The tasks above are basically instance-level or category-level. To prove the effectiveness of our method on segmenting images at any spatial granularity, we evaluate \MethodName on the large-scale part segmentation benchmark PartImageNet~\cite{partimagenet} and the results are shown in \Cref{tab:ps}.
\begin{table}
    \centering
    \small
    \renewcommand\arraystretch{1.1}
    \caption{Results on part segmentation benchmark.}
    \setlength\tabcolsep{18.2pt}
    \label{tab:ps}
    \begin{tabular}{lccc}
    \hline
    Method  &Val IoU & Test IoU   \\
    \hline
       SemanticFPN~\cite{pfpn} &56.76 & 54.57 \\
       DeepLabV3+~\cite{deeplabv3}  &60.57 & 58.71   \\
       SegFormer~\cite{seggpt} &61.97 & 61.46  \\
       \hline
       \MethodName-20 &62.46 &62.03 \\
       \MethodName-100 &\textbf{63.87} &\textbf{63.62} \\
       \hline
    \end{tabular}
    \vspace{-5pt}
\end{table}

\subsection{Training Source Component}

\begin{figure}[t]
    \centering
    \includegraphics[width=\linewidth]{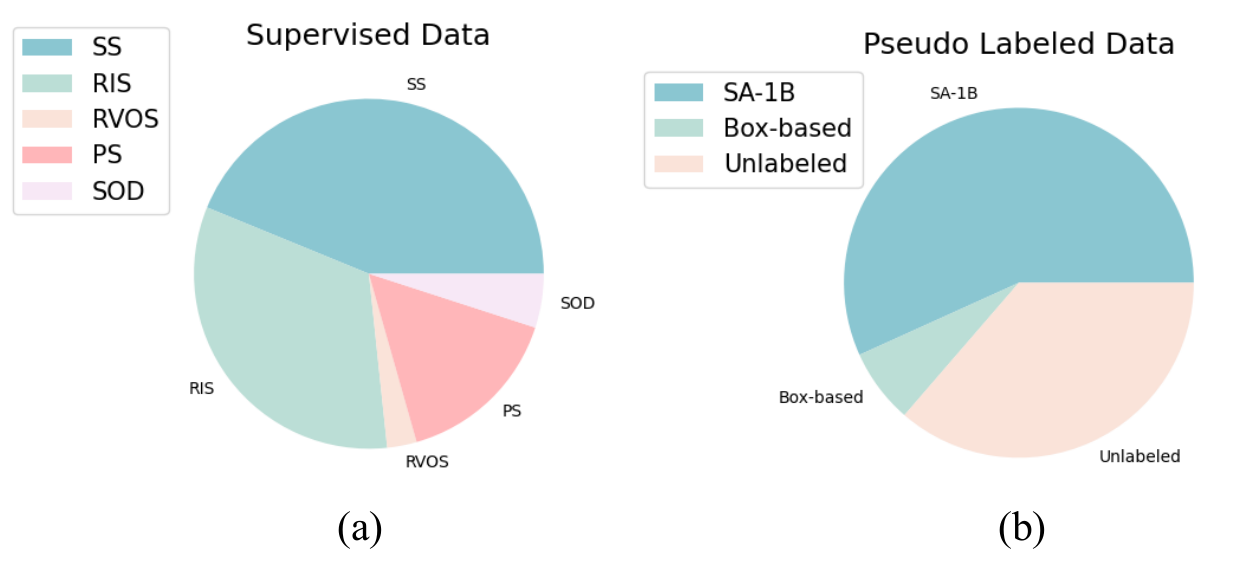}
    \caption{Illustration of training data component. (a) shows the proportions of supervised  source collected from different tasks. (b) demonstrates the component of pseudo labeled training source.}
    \label{fig:data_component}
\end{figure}

In this part we analyze the composition of training data.
\Cref{fig:data_component} (a) shows the proportions of supervised data collected from different tasks. 
From the linguistic perspective, language expressions from RIS and RVOS are natural long linguistic captions. That from other tasks are coarse-grained short expressions, \textit{e.g.}, category names. 
From the visual standpoint, RIS, RVOS, and SOD are instance-level understanding. SS and PS are about semantic-level (scene-level) and local part-level, respectively.
The total number of supervised images and mask-caption pairs is 360k and 7.58M.

\Cref{fig:data_component} (b) demonstrates the component of pseudo labeled training source generated from weakly annotated and unlabeled data by the automatic annotation engine. ``SA-1B" is corresponding to the mask-based data. The figure present the statistic for 20\% SA-1B. Under such circumstances, we totally collect 3.5M images with 22M pseudo mask-caption pairs. For 100\%  SA-1B data, the total number is about 11.5M images with 126M mask-caption pairs.

\subsection{Ablation Study}

\paragraph{Architecture Analysis}
To verify that the model design of our \MethodName is compatible with the language prompt paradigm, we perform ablation experiments on different parts of the network and the results are shown in \Cref{tab:model_analysis}. PF, LP, and VP indicate Pre-Fusion, Vision Path, and Language Path, respectively. We take RIS, SS, and SOD tasks for evaluation. It is evident that the model's performance on the pertinent tasks enhances as the synergy between language and visual content intensifies.

\begin{table}[t]
    \centering
    \small
    \setlength\tabcolsep{10.5pt}
    \renewcommand\arraystretch{1.1}
    \caption{Ablation studies about the model components. The experiments are performed on RIS, SS, and SOD tasks with RefCOCO, ADE20k and SOD benchmarks, respectively.}
    \begin{tabular}{l|c|c|c}
        \hline
        Method & RefCOCO & ADE20K & SOD\\
        \hline
        Baseline &48.68 &21.89 &0.671\\
        +PF &65.38 &37.45 &0.784\\
        +PF, +LP &74.47 &43.12 &0.811\\
        +PF, +LP, +VP &\textbf{77.56} &\textbf{44.63} &\textbf{0.827} \\
        \hline
    \end{tabular}
    \label{tab:model_analysis}
    \vspace{-5pt}
\end{table}

\begin{figure*}[t]
    \centering
    \includegraphics[width=\textwidth]{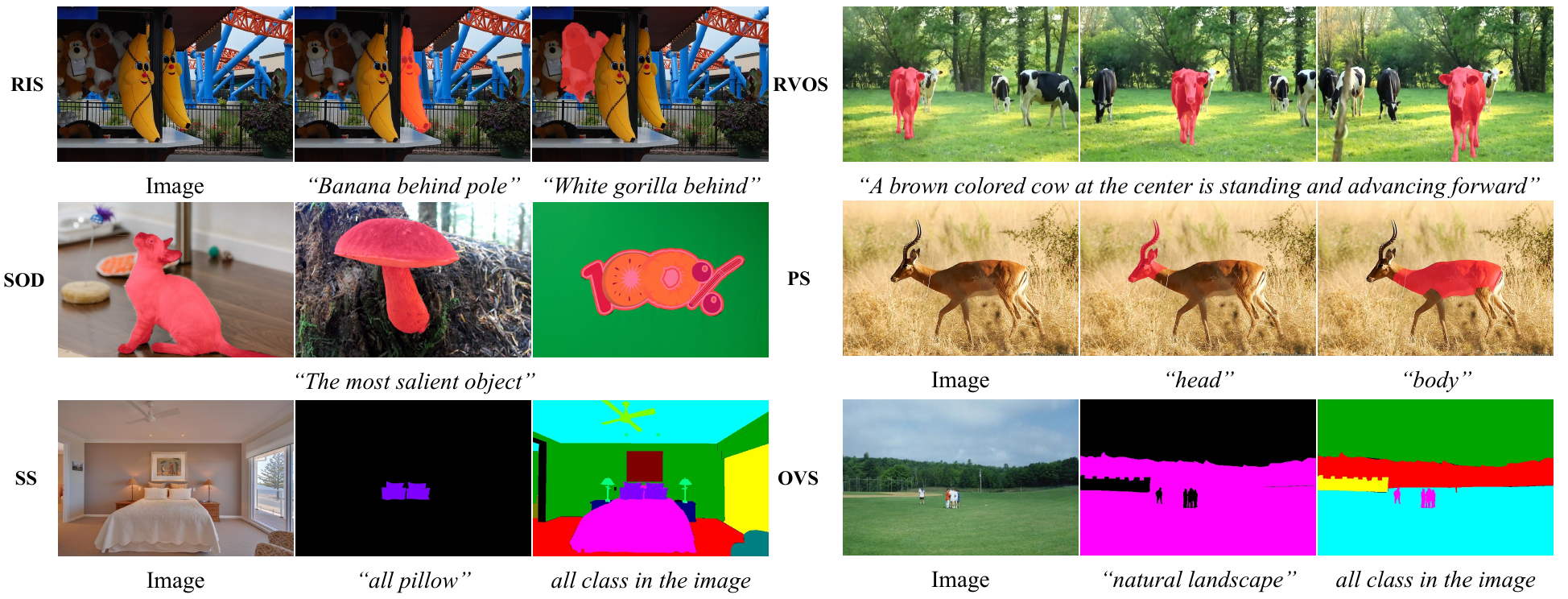}
    \vspace{-15pt}
    \caption{Visualization of segmentation results for different tasks.}
    \label{fig:results}
    \vspace{-5pt}
\end{figure*}

\vspace{5pt}
\noindent \textbf{Training Source Analysis}
This part shows the impact of different training source on segmentation performance. 

\begin{figure}[t]
    \centering
    \includegraphics[width=0.85\linewidth]{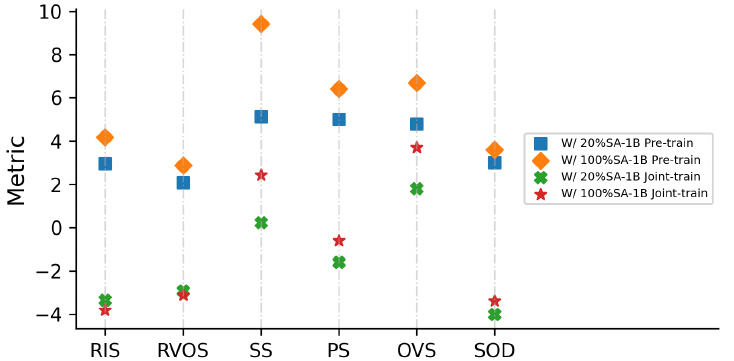}
    \caption{Effect of incorporating 20\% as well as 100\% SA-1B data into training process under pre-training and joint training strategy.}
    \label{fig:sa1b}
    \vspace{-15pt}
\end{figure}

\textit{Effectiveness of SA-1B data:}
SA-1B~\cite{sam} contains a large number of high quality images from diverse scenarios. However, due to the lack of textual labels, we need to leverage existing models to generate pseudo mask-caption pairs, which leads to annotation noise and attenuates its effect. We have tried two strategies, \ie, joint training and pre-training, for incorporating it into the training process. 
\Cref{fig:sa1b} shows the effect of sampling 20\% and 100\% SA-1B data into training process under these two strategies. The vertical axis represents the performance increase or decrease compared to training without SA-1B data. It can be seen that the pre-training strategy performs significantly superior to joint training. We attribute this to the presence of heavy noise in the pseudo-labeled SA-1B data disrupting the normal training space. In addition, 100\% SA-1B is significantly better than 20\%  under pre-training, but the phenomenon is not the same under joint training due to the larger noise distribution.


\begin{figure}[t]
    \centering
    \includegraphics[width=\linewidth]{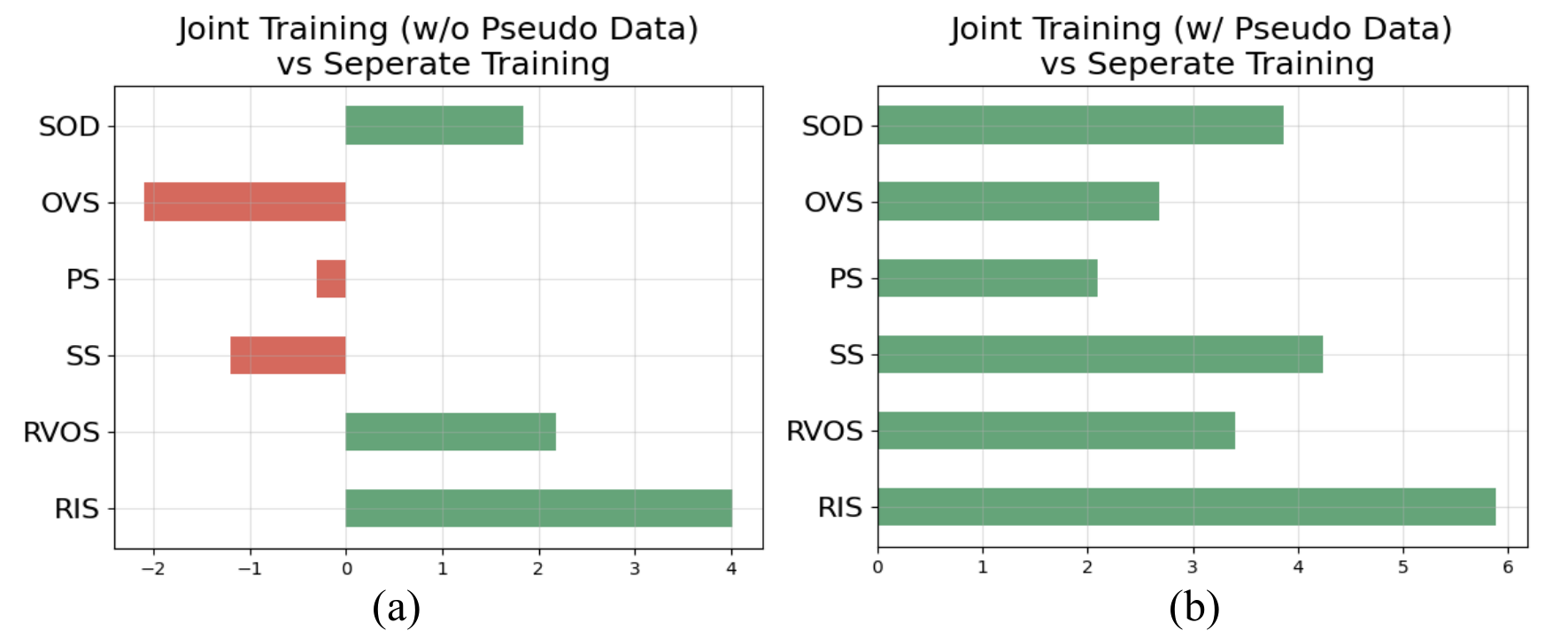}
    \vspace{-15pt}
    \caption{Influence of multi-task joint training. (a) shows the results without large-scale pseudo-labeled data for pre-training. (b) demonstrates results trained with pseudo-labeled data.}
    \label{fig:multi-task}
    \vspace{-10pt}
\end{figure}

\textit{Effectiveness of multi-task joint training:}
We also test the performance gains resulting from multi-task joint training. From \Cref{fig:multi-task} (a) we can see that vanilla multi-task joint training does not result in performance boost on all tasks without pseudo-labeled data aiding training.
For short text caption tasks, multi-task joint training leads to decrease. 
We believe this is due to differences in visual and linguistic distribution across tasks.
While with the large-scale pseudo-labeled data for pretraining, this problem can be greatly mitigated by superior initialization, as shown in \Cref{fig:multi-task} (b).

\subsection{Visualization Results}
\Cref{fig:results} shows some segmentation examples for each task, which demonstrates the capability of our model to segment images at arbitrary semantic granularity with language instructions. Due to limited space, please see more results in supplementary materials.

\section{Conclusion}
In this paper we aim to achieve universal segmentation of arbitrary semantic level with language instruction. 
We reorganize a group of tasks from original diverse distributions into a unified data format for joint training, \ie, triplet of images, masks, and captions. 
To promote the model's understanding of high-level language instructions, we present a fully aligned framework called \MethodName. Combined with an automatic annotation engine for leveraging numerous unlabeled or weakly labeled data, our \MethodName achieves superior performance on various semantic-related tasks.

\noindent \textbf{Acknowledgements.} This work was supported in part by the National Natural Science Foundation of China under Grant 62206153 and No. U1903213, in part by Shenzhen Science and Technology Program under Grant CJGJZD20220517142402006 and JCYJ20220818101014030.

{
    \small
    \bibliographystyle{ieeenat_fullname}
    \bibliography{main}
}


\end{document}